\documentclass[10pt,journal,compsoc]{IEEEtran}
\ifCLASSOPTIONcompsoc
  \usepackage[nocompress]{cite}
\else
  \usepackage{cite}
\fi
\usepackage{amsmath,amssymb}
\usepackage{diagbox}
\usepackage{multirow}
\usepackage{booktabs}       % professional-quality tables
\usepackage[switch]{lineno} %-

\def\etal{\emph{et al.\ }}

\ifCLASSINFOpdf
   \usepackage[pdftex]{graphicx}
   \graphicspath{{../}}
\else
\fi
\usepackage{amsmath}
\begin{document}
\title{Dividing and Conquering Cross-Modal Recipe Retrieval: from Nearest Neighbours Baselines to SoTA}

\author{Mikhail~Fain,
        Niall~Twomey,
        Andrey~Ponikar,
        Ryan~Fox,
        and~Danushka~Bollegala% <-this % stops a space
\IEEEcompsocitemizethanks{\IEEEcompsocthanksitem M. Fain, N. Twomey and A. Ponikar are with Cookpad Ltd. R. Fox is with Facebook (his contributions here were made prior to joining Facebook), and D. Bollegala is with the University of Liverpool.\protect\\
E-mail: andrey-ponikar@cookpad.com}}% <-this % stops an unwanted space

\IEEEtitleabstractindextext{%
\begin{abstract}

We propose a novel non-parametric method for cross-modal recipe retrieval which is applied on top of precomputed image and text embeddings. By combining our method with standard approaches for building image and text encoders, trained independently with a self-supervised classification objective, we create a baseline model which outperforms most existing methods on a challenging image-to-recipe task. We also use our method for comparing image and text encoders trained using different modern approaches, thus addressing the issues hindering the development of novel methods for cross-modal recipe retrieval. We demonstrate how to use the insights from model comparison and extend our baseline model with standard triplet loss that improves state-of-the-art on the Recipe1M dataset by a large margin, while using only precomputed features and with much less complexity than existing methods. Further, our approach readily generalizes beyond recipe retrieval to other challenging domains, achieving state-of-the-art performance on \textsc{Politics} and \textsc{GoodNews} cross-modal retrieval tasks.
    
\end{abstract}

% \begin{abstract}
%   Non-literal cross-modal retrieval tasks concern the domains where the two modalities are complementary rather than expository. We propose a novel non-parametric method for cross-modal retrieval which is applied on top of precomputed image and text embeddings. By combining our method with standard approaches for building image and text encoders, trained independently with a self-supervised classification objective, we create a baseline model which is competitive with advanced approaches on three different non-literal cross-modal tasks. We further demonstrate that our method is well-suited for comparing image and text encoders trained using different approaches on a canonical example of a non-literal task, cross-modal recipe retrieval, thus addressing the issues hindering the development of novel methods in this domain. We show how to use the insights from model comparison to go beyond the baselines and achieve SoTA on the Recipe1M dataset by a large margin, while using only precomputed features and with much less complexity than existing methods.
% \end{abstract}

\begin{IEEEkeywords}
Cross-modal retrieval, baselines, nearest neighbours.
\end{IEEEkeywords}}

\maketitle

\IEEEdisplaynontitleabstractindextext

% \fi
\IEEEpeerreviewmaketitle

\IEEEraisesectionheading{\section{Introduction}\label{section:introduction}}

% \IEEEPARstart{W}{e} consider the problem of developing baselines for cross-modal retrieval tasks. Cross-modal retrieval is a well-researched topic \cite{tkde_crossmodal}, and boasts definitive tasks and datasets that are investigated by much of the published research, including Flickr30k \cite{plummer2015flickr30k} and MS-COCO \cite{lin2014microsoft}. These datasets are characterised by a literal correspondence between a short textual caption and the  image, e.g. an image corresponding to the caption of `ball' will surely contain a ball.

\IEEEPARstart{I}{n} this work we are exploring the problem of cross-modal recipe retrieval between food images and textual cooking recipes. A solution to this problem has a number of applications, such as searching for the correct recipe using a photo \cite{Marin2018}, automatically determining the number of calories in a dish \cite{Myers2015} and improving the performance of various recipe recommendation and ranking systems \cite{Freyne}. 
% NT (a bit confusing at this stage) Note, that we primarily focus on image-to-textual-recipe retrieval rather than textual-recipe-to-image retrieval for the sake of clarity. 
This task involves searching for an exact matching text of the recipe given its image among candidate textual recipes from a held-out test set. 

The problem of recipe retrieval is challenging due to the diverse nature of food images and the subtle differences between recipes. For example, the photos of dishes made by following the same recipe could look completely different from each other, and very similar dish photos could be associated with very different ingredients and procedures\footnote{As an example, consider visually distinguishing different types of creamy soups from each other using only photos}.

% On the other hand, there exist other scenarios where the relationship between the text and the image is more nuanced. Take the example where a caption refers to a particular legal case in constitutional law that is under review with a corresponding image depicting `Lady Justice'. It is clear that while the text and image pairing is suitable, the relationship between the two is rather more subtler than the previous example with the ball. Following the recent nomenclature of Thomas and Kovashka \cite{thomas2020preserving}, we denote the datasets primarily featuring this more subtle relationship as \textit{non-literal} (while the Flickr30k and MS-COCO tasks are termed \textit{literal}). 
% We observe that approaches to modelling non-literal tasks are often characterized by specialized methods due to structure of the data \cite{Salvadora,thomas2020preserving}, and such problems received less research attention from the community owing to the recent emergence of their definition. This motivates our focus on finding robust and efficient baselines particularly for this type of datasets to facilitate future progress. 

After the release of the \textsc{Recipe1M} dataset \cite{Salvadora} containing diverse food images and recipes, cross-modal recipe retrieval became one of the standard benchmarks for image-to-document retrieval tasks, with numerous improvements made on the original model \cite{Salvadora}. Despite this progress, we observe a few issues that, if not addressed, may hinder our understanding of added value delivered by new methods. Namely, these issues are: the lack of strong baselines and the complexity of identifying strengths and weaknesses of individual model components across methods.

% A prime example of a large-scale non-literal task is cross-modal recipe retrieval \cite{Marin2018}, since the details of the cooking recipe often get reflected in the image only in subtle and indirect ways (e.g., the verb \textit{fry} appearing in the recipe could be weakly correlated with more crispy-looking dish photos). After the release of the \textsc{Recipe1M} dataset \cite{Salvadora} containing diverse food images and recipes, cross-modal recipe retrieval became the definitive specialized image-to-document retrieval task.
% % NT: not necessary - discussed in the problem definitions... , with numerous improvements made on the original model \cite{Salvadora}. 
% % Despite this progress, we observe a few issues hindering further development of the new methods in this space: 
% Despite this progress, we are wary of a few issues that, if not addressed, may hinder our understanding of added value delivered by new methods: 
% % NT namely, the lack of strong baselines and the complexity of identifying strengths and weaknesses of individual model components across methods. 
% % NT (haven't discussed our work yet, seems premature) We note, that we only focus on image-to-text retrieval in this work rather than text-to-image retrieval for the sake of clarity.

\textsc{Problem 1}: \textit{lack of strong baselines}.
In the seminal paper on cross-modal recipe retrieval, Salvador \etal \cite{Salvadora} presented a baseline model using a classic 
% NT: one of the strongest 
statistical models for learning joint embeddings, Canonical Correlation Analysis (CCA) \cite{cca}. This model achieved the \mbox{top-1} accuracy of 14.0 on a test set of 1,000 recipes. A deep learning model presented in the same paper almost doubled the top-1 accuracy reaching 25.6. By 2019, after two years of steady improvements \cite{Chen2017crossmodal,Chen2018crossmodal,Carvalho2018,Zhu_2019_CVPR}, this metric was doubled again to reach 51.8 \cite{wang2019learning}. While this could reflect the improvements in retrieval methods, it could also be due to mis-specified baselines in this domain. 
% MF: The rapid and significant advance of performance suggests that CCA cannot be considered as a strong baseline anymore, highlighting the need for acceptance of new baselines. 
If the latter is the case, a strong and simple baseline for cross-modal retrieval could facilitate faster scientific progress and put the previously reported results into perspective. 

\textsc{Problem 2}: \textit{challenges in identifying model's components strengths}. 
Cross-modal recipe retrieval models are very complex and have many interacting parts \cite{wang2019learning}. Thus the performance gains are not easy to attribute to improvements in a particular component (e.g. the image or text encoder). This makes it extremely hard to understand and rank the various elements in the model, as ablation studies only address performance gains within the same method. The prior work on cross-modal recipe retrieval \cite{Chen2018crossmodal,Carvalho2018,Zhu_2019_CVPR,wang2019learning} mitigates this issue by using an image and textual processing pipeline that is similar to the one introduced by Salvador \etal \cite{Salvadora}. However, the setups still sufficiently differ to prevent a fair comparison \cite{Chen2018crossmodal,Carvalho2018}, and, in addition, being tied to specific architectures may limit the scope of future research. 
% presents a danger of the new methods being tailored to them and makes it especially hard to drive improvements in encoding images and text. 

We introduce Cross-Modal k-Nearest-Neighbours (CkNN) in Section \ref{section:method} to address the aforementioned problems. We use k-Nearest-Neighbours (kNN) to search over modalities using the correspondences available in the training set. Our results are presented in Section \ref{section:results}: specifically, we apply CkNN to challenging non-literal image-to-text retrieval tasks by leveraging standard approaches for independently representing images and text using a self-supervised classification objective. This is a solution for \textsc{Problem 1} since we define a straightforward competitive baseline on cross-modal recipe retrieval task. Since CkNN depends directly on distance measures across different modalities, we use it for comparing the efficacy of image and text encoders, addressing \textsc{Problem 2}. We demonstrate how to use the insights from encoder comparison and go beyond our baseline results to improve state-of-the-art (SoTA) by a large margin on \textsc{Recipe1M} while still using only precomputed features and standard approaches. We further show that our approach generalizes to other challenging cross-modal retrieval tasks, reaching SoTA performance on \textsc{Politics} \cite{thomas2019predicting} and \textsc{GoodNews} \cite{biten2019good} datasets. 

The combination of our contributions raises the bar for baseline models and model component analysis in cross-modal retrieval and sets a new SoTA reference performance on \textsc{Recipe1M}. We hope that our approach would encourage further development of advanced end-to-end methods.

\section{Related Work} \label{section:related_work}

The problem of cross-modal retrieval has been researched extensively in the Computer Vision community with a primary focus on datasets where there exists a clear mapping between objects in the image scene and a concise textual description of the image  \cite{thomas2020preserving}. These tasks are exemplified by the standard \textsc{Flickr30k} \cite{plummer2015flickr30k} and \mbox{MS-COCO} \cite{lin2014microsoft} benchmarks. The majority of the solutions create separate representations for the two different modalities, projecting them to the same shared space and performing a similarity search within that space \cite{jiang2017,tkde_wang2020}. Such models are typically based on neural networks and are trained end-to-end \cite{Feng2014,Ngiam2011,tkde_tu2020}. The recently proposed methods exploited semantic category labels to learn discriminative features for cross-modal retrieval \cite{Peng2017CCL,Wang2017,tkde_shen2020}. Adversarial learning \cite{gan} has also been employed to aid cross-modal retrieval \cite{Peng2017}. Further work shows the benefits of applying attention on top of the object detection pipeline to capture fine-grained relationships between vision and language, creating a better aligned joint embedding space \cite{Lee2018}. Wang \etal \cite{wang2019camp} extend these approaches to move away from the shared embedding space completely.

Some of the above ideas could be adapted successfully to cross-modal \textit{recipe} retrieval task, which is a subproblem of the general cross-modal retrieval problem featuring a subtle image-text relationship. In this case, one modality is a recipe image, and the second one is a structured text, consisting of a recipe title, a list of ingredients in free form and a list of instructions, also written in free form. It was introduced by Salvador \etal \cite{Salvadora}, who used margin loss for learning the shared embedding space. The image processing pipeline was based on ResNet-50 \cite{resnet50}. Recipe ingredients were normalized using a separate model involving bi-directional LSTM \cite{hochreiter1997} and further encoded with word2vec \cite{mikolov2013}. The list of instructions was encoded using skip-thoughts \cite{kiros2015}. The encoded ingredients and instructions were then passed through to separate LSTMs and concatenated, thus generating the encoding of the recipe text. The resulting model was trained end-to-end (except for word2vec and skip-thoughts vectors which were pretrained separately), improving the top-1 accuracy of \textbf{CCA} \cite{cca} baseline from 14.0 to 25.6 on a test set of size 1,000 \cite{Salvadora}. This model is further referred to as Pic2Recipe.

The follow-up work has been largely focused on expanding on the above setup, with the focus on improving cross-modal alignment techniques and minor changes in individual modality processing pipelines and training methods. For example, Chen \etal \cite{Chen2017crossmodal,chen2017crossmodal_b} analyzed the importance of instructions and ingredients for cross-modal retrieval and then built a text representation designed to match Pic2Recipe performance \cite{Chen2018crossmodal}, despite relying on a Convolutional Neural Network (CNN) pretrained on another food image dataset rather than learning it end-to-end. This model is denoted in this paper as AM following \cite{wang2019learning}. 

Carvalho \etal \cite{Carvalho2018} made improvements to the alignment loss function using double-triplet loss in their AdaMine model, improving the top-1 accuracy to 39.8. Zhu \etal \cite{Zhu_2019_CVPR} in their R2GAN method employed Generative Adversarial Networks (GANs) \cite{gan} to help with learning the representations and reaching results similar to AdaMine. MCEN method uses stochastic latent variable model to share the information between modalities \cite{Fu_2020_CVPR}. The current SoTA, ACME \cite{wang2019learning}, also uses GANs in addition to a cross-modal triplet loss scheme \cite{Wang2017} together with an effective sampling strategy \cite{Schroff2015}, modality alignment using an adversarial learning strategy from \cite{Wang2017} and a cross-modal translation consistency loss to reach an impressive 51.8 top-1 accuracy, more than doubling the original performance of Pic2Recipe. We note that we were unable to reproduce the reported performance of ACME model with the model weights released by the authors, and refer to the model achieving the slightly lower score as ACME*.

Most of the described cross-modal recipe retrieval models reported significant benefits from using a \textit{semantic regularization} technique \cite{Salvadora}, where the image and text embeddings are constrained by an additional classification loss with the labels being the categories of the recipes.

On the related topic of building powerful food image classifiers there has been an independent body of work focused around food image datasets \cite{Bossard,Chen2017,Hassannejad2016,Singla2016}. The researchers have explored a variety of architectures more suitable to food images than the ResNet-50 backbone used in cross-modal retrieval \cite{food101sota}.

For recipe text encoding, the body of literature is less organized. As there are no commonly used benchmarks for evaluating recipe text representations, the models are usually tuned as part of a bigger task, such as cross-modal retrieval \cite{Salvadora}, recipe translation \cite{sato-etal-2016-japanese} or ingredient pairing \cite{Donghyeon2019}.

\begin{figure*}[t]
    \centering
    \includegraphics[width=0.7\textwidth]{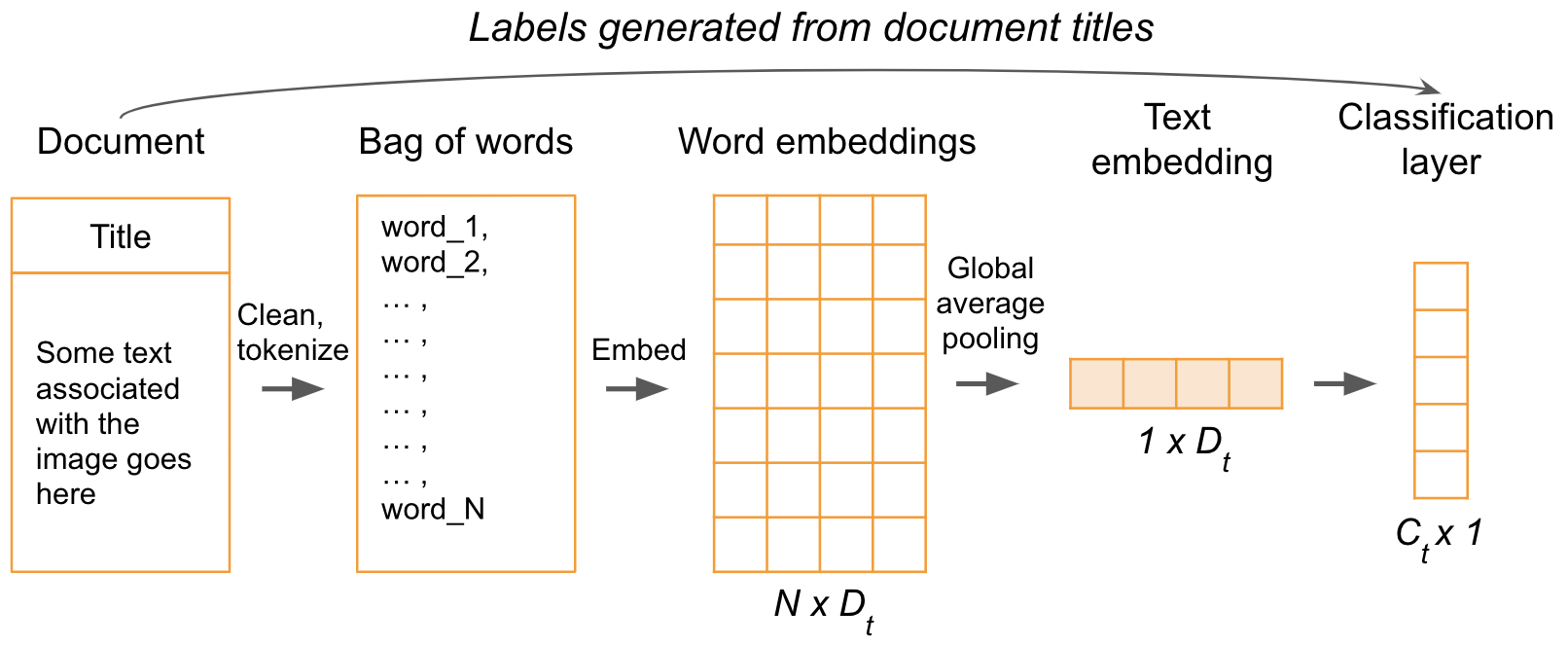}
    \caption{Average word embeddings encoder.}
    \label{fig:bow}
\end{figure*}

In addition to recipe retrieval, there exist other cross-modal retrieval domains where the relationship between text and images are more subtle than the standard benchmarks like Flickr30k and MS-COCO. Thomas and Kovashka \cite{thomas2020preserving} name these types of tasks as \textit{non-literal} cross-modal retrieval and show that enforcing neighbourhood consistency between the image and text spaces during training is beneficial to performance, with their best model denoted SN outperforming the baselines on \textsc{Politics} \cite{thomas2019predicting} and \textsc{GoodNews} \cite{biten2019good} datasets.

Nearest Neighbour Search has been shown to outperform many more complex deep learning methods on neural recommendation tasks \cite{Dacrema:2019aa}. It has also been used successfully for fine-grained image retrieval as a base for many query expansion \cite{Ondrej2007} and database augmentation \cite{Panu2009} strategies, which work on top of other methods and yield significant gains on image retrieval tasks. 
\footnote{https://landmarksworkshop.github.io/CVPRW2019/}. 
However, to the best of our knowledge, these ideas have not been extended for cross-modal retrieval.

\section{Method} 
\label{section:method}

Since the main purpose of this work is to create strong baselines and compare encoders rather than aim for the best possible method of cross-modal retrieval, we do not train our model end-to-end unlike existing approaches \cite{Salvadora,Carvalho2018,wang2019learning}. 
% Instead we build a text encoder (Section \ref{section:awe_encoder}), an image encoder (Section \ref{section:recipe1m_classifier}) and an alignment module (Section \ref{section:cknn}) independently. 
Instead, we adopt a widely-used approach for independently training encoders using a self-supervised classification objective \cite{Bert,Doersch2015}, which we setup by automatically extracting noisy labels from text. Our encoders employ basic architectures: the last layer of a CNN for images \cite{RetrievalCNN,Faster_R_CNN} and the average of word embeddings (AWE) for text. \cite{Wieting2016,Le2014,Hill2016LearningDR}. We describe the self-supervised task for the text and image modalities respectively in Sections \ref{section:awe_encoder} and \ref{section:recipe1m_classifier}, and Section \ref{section:cknn} introduces the manner in which they are integrated. 

% We adopt a widely-used approach for independently training encoders using a self-supervised classification objective \cite{Bert,Doersch2015}, which we setup by automatically extracting noisy labels from text. We describe the self-supervised task for the text and image modalities respectively in Sections \ref{section:awe_encoder} and \ref{section:recipe1m_classifier}, and Section \ref{section:cknn} introduces the manner in which they are integrated. 

\subsection{Average Word Embeddings Encoder}
\label{section:awe_encoder}

We use a standard average word embeddings \cite{Wieting2016,Le2014,Hill2016LearningDR,tkde_guo2019} model as a text encoder. This model assigns an embedding in $D_{t}$ dimensions to each word in the input document, and computes the document representation by averaging the embeddings. Note, that this is one of the most basic approaches in the literature with the structure of the document being completely ignored by such representation.

To train the word embeddings, we use text data to define a self-supervised task. We follow the approach of \cite{Salvadora} and extract a set of $C_t$ noisy labels based on frequent unigrams and bigrams from document title and filter them by a threshold. The remaining text is treated as features which are used to learn a mapping to the label space.  Therefore, we use the main body (ingredients and instruction) of the textual recipe as training input data, and the title to extract labels.
% To build a text encoder for a long document which can be split into a title and main body parts, we train a classifier which predicts automatically extracted labels. To extract a set of $C_t$ noisy labels, we select frequent unigrams and bigrams from the title inspired by \cite{Salvadora}, allowing overlapping labels and multiple labels per document. 
% Therefore, we use the main body of the textual document as training input data, and the title to extract labels. 
% 

The details of our training setup could be described as follows. The encoder input is treated as one long, continuous document. We then add a trainable, randomly initialized word embedding layer with $D_{t}=300$ dimensions. The next layer computes the mean of all the embeddings, a linear layer with sigmoid activations is added on top for multi-label classification. Binary cross-entropy is used as a loss function. See full architecture in \figurename \ref{fig:bow}. We refer to this model as the \textit{AWE-Encoder}. 

While this self-supervised task is applicable for textual recipes, it does not apply to datasets where the documents do not have identifiable title (such as in \textsc{GoodNews} dataset we explore in Section \ref{section:other}, which consists of short captions paired with images). In this case, we train word embeddings using unsupervised FastText \cite{bojanowski2016enriching} method instead of the classification objective.

% When inferring the text embedding, we take the average of word embeddings for the text document.

\subsection{Image Encoder} \label{section:recipe1m_classifier}

% For creating an image encoder we train a multi-label classifier on images with a set of noisy labels extracted from the text or its part (e.g., title) in the same way as for AWE-Encoder.
Similar to AWE-Encoder, our image encoder is trained using the set of $C_i$ noisy labels extracted from text. Image representations are produced using ResNet-50 \cite{resnet50}, and we apply binary cross-entropy loss for multi-label classification. To create the image embeddings, we extract features from the last convolutional layer of the network \cite{RetrievalCNN,Faster_R_CNN} after global average pooling. We refer to this model as \textit{ResNet-Encoder}.

\subsection{Cross-Modal k-Nearest-Neighbours (CkNN)} \label{section:cknn}

\begin{figure*}[t]
    \centering
    \includegraphics[width=\textwidth]{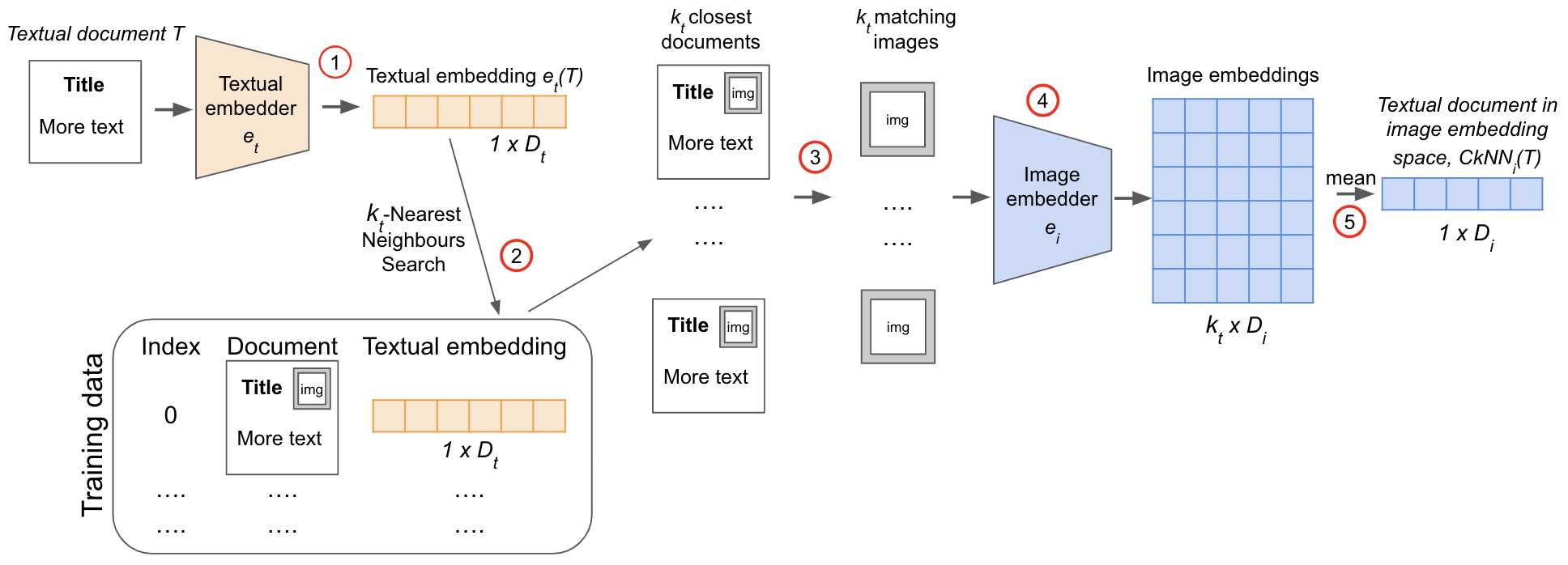}
    \caption{A schematic representation of the core idea of our CkNN method: representing a text $T$ in the image embedding space using nearest neighbour search in the training data. Orange and blue shapes respectively denote text and image modalities. Numbers in red circles correspond to $\mathrm{CkNN}_{i}(T)$ algorithm steps described in Section \ref{section:cknn}}
    \label{fig:cknn}
\end{figure*}

CkNN belongs to the category of alignment modules, which attempt to match the representations of different modalities. It is applied on top of an image encoder $e_{i}$ with a distance measure $d_i(\cdot, \cdot)$ in the image embedding space, and a text encoder $e_{t}$ with a distance measure $d_t(\cdot, \cdot)$ in the text embedding space. In this work we use cosine similarity as a distance measure for both the text and image embedding spaces.
% \todo{talking about distance here is surprising - why is it needed?}In practice, we use cosine distance in the embedding space as the distance measures $d_{i}$ and $d_{t}$, although they could be different from each other in principle. 
%Since our methodology relies on kNN, we measure the distance between texts (and images) using cosine similarity ($d_t(\cdot, \cdot)$ and $d_i(\cdot, \cdot)$ respectively). Alternative distance measures may also be considered. 

CkNN, depicted in \figurename \ref{fig:cknn}, uses the training data to represent a candidate text $T$ in the image embedding space using the following algorithm denoted as $\mathrm{CkNN}_{i}(T)$: 
% CkNN 

\begin{enumerate}
    \item{Encode a candidate text document $T$ using $e_{t}(T)$.}
    \item{Find the $k_t$ nearest neighbours based on the text embeddings using $d_{t}$, denoted $\mathcal{R}_{T}$.}
    \item{Extract the set of images $\mathcal{I}_T$ associated with $\mathcal{R}_{T}$.}
    \item{Encode each $I \in \mathcal{I}_T$ in the image embedding space using $e_i(I)$.}
    \item{Return the mean vector $\frac{1}{|\mathcal{I}_T|}\sum_{I \in\mathcal{I}_T}e_i(I)$.}
\end{enumerate}

\begin{figure}[t]
    \centering
    \includegraphics[width=1.0\columnwidth]{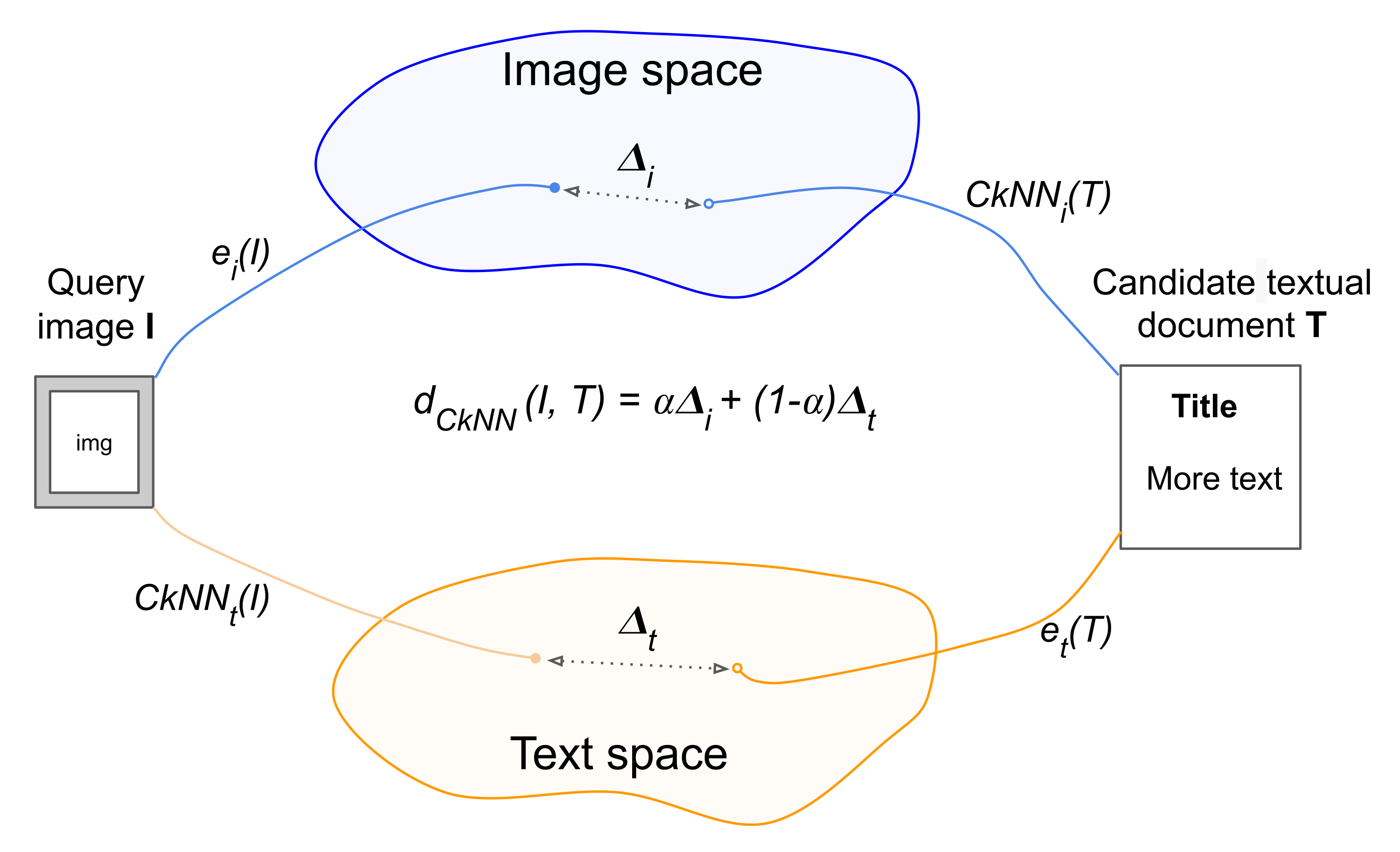}
    \caption{CkNN distance $d_{\mathrm{CkNN}}(I,T)$ between a query image $I$ and a candidate textual recipe $T$ to be used in ranking for the cross-modal retrieval task}
    \label{fig:cknn_2}
\end{figure}

The complementary algorithm ($\mathrm{CkNN}_{t}(I)$) produces text embeddings for an image query, $I$, allowing a separate neighbourhood size, $k_{i}$.
Thus, we now have two ways to calculate the distance between a query image $I$ and a candidate text $T$: (a) in the image embedding space, and (b) in the text embedding space. We define the total distance between an image and a document as the linear combination of the two shown in Eq.\ (\ref{eq:cknn}) and illustrated in \figurename \ref{fig:cknn_2}. This distance can be used to rank the set of candidates based on a particular query. 

\begin{equation}
\label{eq:cknn}
\begin{split}
    d_{\mathrm{CkNN}}(I, T) = & d_{i}(e_i(I), \mathrm{CkNN}_{i}(T)) \, \alpha + \\
    & d_{t}(\mathrm{CkNN}_{t}(I), e_t(T)) \, (1-\alpha) 
\end{split}
\end{equation}

We emphasize that we developed the non-parametric \textsc{CkNN} method for the purpose of creating cross-modal retrieval baselines rather than achieving state-of-the-art results. Further, the encoders from Sections \ref{section:recipe1m_classifier} and \ref{section:awe_encoder} are trained using basic methods, and the resulting embeddings are not designed to be used for retrieval using cosine similarity. We refrain from applying more advanced methods for representation learning \cite{Bert} as well as metric learning \cite{Deng2018ArcFaceAA} for our modelling components to keep the focus on baselines and standard approaches.

\section{Experiments and Results} \label{section:results}

\begin{table*}[t]
\caption{Performance of our baseline method and other reported methods on \textsc{Recipe1M} image-to-textual-recipe task. The best results are shown in bold. The description of the existing methods could be found in Section \ref{section:related_work}. The metrics for the models denoted by \textit{*} are computed by us using publicly available models. 
%$^\dag$ symbol indicates that the model was trained as part of our ablation study. 
While we report the results for both 1K and 10K test sample size, we only rely on 10K values for our analysis since 1K results are too noisy \cite{Carvalho2018}}
\label{table:results1}
\begin{center}
\begin{tabular}{c l rrrr}  
\toprule
Size of test set & Method & medR $\downarrow$ & R@1 $\uparrow$ & R@5 $\uparrow$ & R@10 $\uparrow$ \\
\midrule
\multirow{12}{*}{1K} & CCA \cite{cca} & 15.7 & 14.0 & 32.0 & 43.0 \\
& Pic2Recipe \cite{Salvadora} & 5.2 & 25.6 & 51.0 & 65.0 \\
& AM \cite{Chen2018crossmodal} & 4.6 & 25.6 & 53.7 & 66.9 \\
& AdaMine \cite{Carvalho2018} & 2.0 & 39.8 & 69.0 & 77.4 \\
& R2GAN \cite{Zhu_2019_CVPR} & 2.0 & 39.1 & 71.0 & 81.7 \\
& \textbf{ACME \cite{wang2019learning}} & \textbf{1.0} & \textbf{51.8} & \textbf{80.2} & \textbf{87.5}  \\
& ACME* \cite{wang2019learning} & 1.8 & 49.0 & 77.1 & 85.2 \\
& MCEN \cite{Fu_2020_CVPR} & 2.0 & 48.2 & 75.8 & 83.6 \\
& (Ours) CkNN+AWE+ResNet & 2.0 & 45.7 & 75.9 & 84.2 \\
\midrule
\multirow{10}{*}{10K} & Pic2Recipe* \cite{Salvadora} & 39 & 7.3 &	20.3 & 29.0 \\
& AM \cite{Chen2018crossmodal} & 39.8 & 7.2 & 19.2 & 27.6 \\
& AdaMine \cite{Carvalho2018} & 13.2 & 14.9 & 35.3 & 45.2 \\
& R2GAN \cite{Zhu_2019_CVPR} & 13.9 & 13.5 & 33.5 & 44.9 \\
& \textbf{ACME \cite{wang2019learning}} & \textbf{6.7} & \textbf{22.9} & \textbf{46.8} & \textbf{57.9} \\
& ACME* \cite{wang2019learning} & 7.5 & 20.6 & 44.3 & 55.7 \\
& MCEN \cite{Fu_2020_CVPR} & 7.2 & 20.3 & 43.3 & 54.4 \\
& (Ours) CkNN+AWE+ResNet & 9.1 & 19.1 & 41.3 &	52.5 \\ 
\bottomrule
\end{tabular}
\end{center}
\end{table*}

% \todo{Needs a couple of references to literal/non-literal concepts}

\subsection{\textsc{Recipe1M} Dataset and Metrics} \label{section:metrics}

The \textsc{Recipe1M} dataset \cite{Salvadora} consists of over 1M textual cooking recipes (including recipe title, list of ingredients in free form, and list of cooking instructions). In addition, 402,760 of those recipes are linked to one or more corresponding images (887,706 images in total). The dataset is split into dedicated training, validation and test sets. 

In our experiments we use the standard metrics on image-to-textual-recipe retrieval task for the \textsc{Recipe1M} dataset as proposed by \cite{Salvadora}:
\begin{itemize}
    \item Recall at Top $K$ (R@$K$) for $K=1,5,10$ describes the percentage of images for which the correct textual recipe is among the top $K$ of the ranked results list (higher is better).
    \item Median Rank (medR) is the median of the rank of the correct textual recipes across all images (lower is better).
\end{itemize}

The cross-modal retrieval metrics are calculated in the same way as in the prior work. Namely, we randomly sample $N=1,000$ (1K setup) or $N=10,000$ (10K setup) recipes from the test set, and for each test image, ranking the textual recipes in the sample using the given model.

The metrics above are noisy due to randomization, so we follow the strategy proposed in \cite{Salvadora} and sample the sets 10 times, reporting the average results. It should be noted that the results are still too noisy in the case of $N=1,000$ due to sampling errors \cite{Carvalho2018}, thus in this work we focus on metrics for $N=10,000$ (although we also report the $N=1,000$ performance for comparison against previously published results in Section \ref{section:sota}).

To focus on one metric for clarity in our evaluation section, we select the recall at rank-1 with the sample size of 10,000 (\textbf{10K-R@1}), but we note that the results are consistent for all of the metrics. Throughout the paper, we only use the Recipe1M validation set for validation purposes, and report the results on the test set.

\subsection{Nearest Neighbour Baselines}
\label{section:competitive}

% \begin{table}[t]
% \caption{Performance of our baseline method and other reported methods on \textsc{Recipe1M} image-to-textual-recipe task for the test set of 10K size.}
% \label{table:results1}
% \begin{center}
% \begin{tabular}{l r}  
% \toprule
% Method & 10K-R@1 \\
% \midrule
% Pic2Recipe \cite{Salvadora} & 7.3 \\
% AM \cite{chen2017crossmodal_b} & 7.2 \\
% AdaMine \cite{Carvalho2018} & 14.9 \\
% R2GAN \cite{Zhu_2019_CVPR} & 13.5 \\
% \textbf{ACME \cite{wang2019learning}} & \textbf{22.9} \\
% ACME* \cite{wang2019learning} & 20.6 \\
% MCEN \cite{Fu_2020_CVPR} & 20.3 \\
% (Ours) CkNN+AWE+ResNet & 19.1 \\
% \bottomrule
% \end{tabular}
% \end{center}
% \end{table}

% For \textsc{Recipe1M}, we have used the recipe title to extract $C_t=3453$ labels for training AWE-Encoder, where we ignored the structure of the rest of the recipe and treated it as a flat document. We used the recipe title and ingredient list to extract $C_i=5036$ labels for training ResNet-Encoder. 
%MF: That's not what we've done for Recipe1M :( Our label set is created by selecting the frequent terms that occur more than $\theta=200$ times in each dataset. 
For \textsc{Recipe1M}, we used the recipe title to extract $C_t=3453$ labels for training AWE-Encoder and treated recipe text as flat documents. We used the recipe title and ingredient list to extract $C_i=5036$ labels for training ResNet-Encoder. 

We then used CkNN to align the modalities. Using the validation split, we found that $\alpha=0.1, \; k_t=15, \; k_i=3 \,$ was suitable in all cases. ResNet-Encoder was trained for 40 epochs with Adam optimizer \cite{kingma2014adam}, a batch size of 512 and an initial learning rate of 0.0001. AWE-Encoder was trained for 15 epochs with Adam optimizer, a batch size of 128 and an initial learning rate of 0.002. The training hyperparameters were manually tuned on validation data.

We report the performance of our proposed baseline method (Section \ref{section:method}) on \textsc{Recipe1M} in  Table \ref{table:results1}. We can see that on \textsc{Recipe1M}, we outperform the majority of the competing methods, even though these models employ sophisticated modelling components. 
%Our results are approximately just 10\% lower than the performance of the SoTA.
%Performance comparison may look to be less striking on the \textsc{Politics} and \textsc{GoodNews} datasets, but the 

%.  owing particularly to their relatively unexplored. However, our performance is still within $\approx 10\%$ of theirs, despite the simplicity of our model. 

% It is unusual to see studies that against other datasets when modelling \textsc{Recipe1M}. 
%Since \textsc{Recipe1M} is considered a canonical dataset for cross-modal retrieval and models are often specialized to the recipe stricture, it is often the sole dataset considered in research papers. Thus, it was surprising to us that our approach, that takes a flat view of documents, is so performant on \textsc{Recipe1M} when compared to historic SoTA performance. Likewise, we view the performance of CkNN model on \textsc{Politics} and \textsc{GoodNews} positively, especially when jointly considering its performance on \textsc{Recipe1M}. We are unaware of any work that evaluates cross-modal retrieval across \textsc{Recipe1M}, \textsc{Politics} and \textsc{GoodNews}, and are thus unable to further discuss the absolute collective performance with respect to existing methodologies. Despite the gap to SoTA, however, we believe there is growing evidence to consider viewing CkNN as a strong baseline in a general sense. 

\subsection{Analysis: Comparison of \textsc{Recipe1M} Image and Text Models} \label{section:comparison}

\begin{table*}[t]
\caption{10K-R@1 metric on the \textsc{Recipe1M} image-to-textual-recipe task, computed for various image and text encoders combined using our CkNN approach (higher is better). All the image models are built on ResNet-50 \cite{resnet50} backbone. Although the results are not optimal, they are competitive with published methods \cite{wang2019learning}, and allow for direct comparisons between different image and textual recipe embeddings.
}
\label{table:results}
\begin{center}
\begin{tabular}{l r r r r r}  
\toprule
\diagbox{Image Model}{Text Model} & ACME* & Pic2Recipe & TF-IDF & AWE-Encoder & Random \\
\midrule
ACME* & 17.9 & 13.3 & 10.6 & 15.6 & 0.01 \\
Pic2Recipe & 8.5 & 7.1  & 5.3 & 7.5 & 0.01 \\
AdaMine & 12.6 & 9.9 & 7.5 & 11.2 & 0.01 \\
ImageNet-Pretrained & 4.4 & 3.4 & 3.8 & 5.0 & 0.01 \\
Food-Pretrained & 7.8 & 6.2 & 6.5 & 8.6 & 0.01 \\
ResNet-Encoder & 16.6 & 13.1 & 12.0 & 17.4 & 0.01 \\
Random & 0.01 & 0.01 & 0.01 & 0.01 & 0.01 \\
\bottomrule
\end{tabular}
\end{center}
\end{table*}

Since CkNN modality alignment fully decouples the image and text encoders and depends directly on image and text distance measures $d_i$ and $d_t$ (Eq. \ref{eq:cknn}), the performance on cross-modal retrieval could indicate how good the individual modality embeddings are for retrieval purposes. As observed in Section \ref{section:competitive}, the performance of CkNN is relatively robust against a wide range of hyperparameter values. This, along with the fact that CkNN does not require training, also contributes to CkNN being a suitable choice for comparing encoders.

We thus compare encoders pretrained with different methods by applying CkNN to all combinations of image and text encoders and report $10K-R@1$ metric on the \textsc{Recipe1M} test set 
\footnote{Here we use only such training recipes for CkNN for which the embeddings are available for all the encoders being compared. This accounts for the small discrepancy with numbers reported in Section \ref{section:sota}.} 
since it is the most studied of the large datasets we consider. For example, we take the image encoder trained jointly as part of SoTA ACME \cite{wang2019learning} model and use it in combination with the text encoder trained jointly as part of Pic2Recipe \cite{Salvadora}. To study this, we first require pre-trained image and text models:

% We observed that the performance of CkNN is robust against a wide range of hyperparameter values, and found the set used in Section \ref{section:competitive} to work well for all the image and text encoders we compare. \todo{Expand on how it is robust, provide metrics and specific cases. Or if this is a preamble, do signposting - as it stands i don't knwo what this section says! }
% This, along with the fact that CkNN does not require training, also contributes to CkNN being a suitable choice for comparing encoders.  % NT: removed since it's re-stating too much IMO

\subsubsection{Image Models}

For comparison purposes, we used the existing image encoders trained for cross-modal recipe retrieval tasks for which we could run the inference code, which are Pic2Recipe \cite{Salvadora}, AdaMine \cite{Carvalho2018} and ACME* \cite{wang2019learning}.

We further used the following baseline image encoders pretrained on different public domain datasets to extract the embeddings from their last convolutional layer: ImageNet-Pretrained pretrained on ImageNet \cite{imagenet_cvpr09} and Food-Pretrained pretrained on the concatenation of Food-101 \cite{Bossard}, ChineseFoodNet \cite{Chen2017} and iFood-2018\footnote{https://github.com/karansikka1/Foodx} datasets. We also used ResNet-Encoder model (Section \ref{section:competitive}), as well as a random embeddings model denoted as {Random}.

\subsubsection{Text Models}

The only two publicly available textual recipe encoders that we could run at the moment of writing are \textbf{ACME*} and \textbf{Pic2Recipe}, described in detail in Section \ref{section:related_work}. Despite our best efforts, we did not manage to run the code for the AdaMine text encoder.

As a baseline unsupervised encoder we represented the textual recipes as a bag-of-subword-units with term frequency-inverse document frequency (TF-IDF) weights calculated on the \textsc{Recipe1M} training set. We applied singular value decomposition on top of this representation, reducing the dimensionality of the embedding to $D_t=2000$. The model is denoted as {TF-IDF}. We also used {AWE-Encoder} model described in Section \ref{section:competitive}, as well as a random embeddings model denoted as {Random}.

\subsubsection{Comparison of Previously Published Model Components}

The results of our experiments are shown in Table \ref{table:results}. The first observation is that the performance of CkNN is competitive with direct search in the embedding space for which the jointly trained encoders were optimized \cite{Salvadora,wang2019learning}. Indeed, combining ACME* image and text encoders using CkNN drops the performance of direct search only from 20.6 to 17.9, which is still better than most previously published methods (Table \ref{table:results1}). For Pic2Recipe, the drop is even smaller: from 7.3 to 7.1.

We further observe that the performance of individual encoders, image or text, is generally consistent across different combinations and in line with performance on the \textsc{Recipe1M} dataset as reported in \cite{wang2019learning}, validating our comparison framework using CkNN. ACME* (SoTA method) image encoder produces consistently higher numbers across all four text encoders than AdaMine (2nd best method) image encoder, and AdaMine is better than Pic2Recipe (3rd best method) across all metrics. ACME* text encoder outperforms Pic2Recipe text encoder across all 6 image encoders.

The two observations above indicate that modality alignment procedures used in the existing approaches add on the order of 10\% improvement to 10K-R@1 metric compared to CkNN. At the same time, the performance differences due to replacing the encoders are around 50\% from the 3rd to the 2nd best result, and from the 2nd best result to the SoTA result. This suggests that the large performance gap in the reported performance of existing cross-modal methods could be explained primarily by the strengths of the learned image and text embedding spaces for retrieval purposes, and not by the quality of cross-modal alignment.

\subsubsection{Comparison of Independently Trained Encoders}

We now compare the results obtained with the encoders trained jointly as part of the existing cross-modal methods and other text and image encoders, trained independently. We observe that the image encoders pretrained on external data combined with unsupervised TF-IDF encoder produce results on a par with some published metrics. Indeed, Food-Pretrained image encoder outperforms Pic2Recipe image encoder in combination with some of the text encoders, and its combination with TF-IDF scores close to direct search with Pic2Recipe. Even Imagenet-Pretrained with TF-IDF produces a reasonable score of 3.8 (a random model would yield 10K-R@1 score of 0.01). 
\footnote{This combination also achieves 1K-R@1 of 15.2 on 1K test set, outperforming CCA baseline of 14.0 reported by \cite{Salvadora}}. 
This shows that CkNN allows easy creation of many competitive baselines out of encoders trained in completely different ways.

Next, we analyze the performance of our proposed self-supervised encoders. Among image encoders, ResNet-Encoder performs close to ACME* and consistently outperforms other encoders. Among text encoders, AWE-Encoder reaches performance similar to ACME*, whereas Pic2Recipe and TF-IDF perform much worse. This suggests that independent training using a self-supervised classification objective can produce encoders competitive with those trained as part of the SoTA cross-modal retrieval methods. In addition, the success of a much simpler average word embeddings architecture compared to the complex ACME* and Pic2Recipe textual model architectures suggests that more research is needed to understand how to best represent textual recipes.

\subsection{Beyond Baselines} 
\label{section:sota}

% \begin{table}[t]
% \caption{Performance of the extensions of our method on \textsc{Recipe1M} image-to-textual-recipe task. The best results are shown in bold, and are statistically significant. The numbers computed by us using publicly available models are denoted by \textit{*}. $^\dag$ symbol indicates that the model was trained as part of our ablation study. Models below mid-line are ours.}
% \label{table:results2}
% \begin{center}
% \begin{tabular}{l rrrr}  
% \toprule
% % Method & medR $\downarrow$ & 10K-R@1 $\uparrow$ & $R_{@5}$ $\uparrow$ & $R_{@10}$ $\uparrow$ \\
% Method & MR  & 10K-R@1  & $R_{@5}$  & $R_{@10}$  \\
% \midrule
% (Previous SoTA) ACME & 6.7 & 22.9 & 46.8 & 57.9 \\
% ACME* & 7.5 & 20.6 & 44.3 & 55.7 \\
% \midrule
% CkNN+AWE+ResNet & 9.1 & 19.1 & 41.3 &	52.5 \\
% CkNN+AWE+ResNext & 6.8 & 22.9 & 46.9 & 58.0 \\
% Triplet+AWE+ResNet & 5.0 & 26.5 & 51.8 & 62.6 \\
% $^\dag$Triplet+AWE+ResNet & 5.9 & 24.4 & 49.4 & 60.5 \\
% \textbf{Triplet+AWE+ResNext } & \textbf{4.0} & \textbf{30.0} & \textbf{56.5} & \textbf{67.0} \\
% $^\dag$Triplet+AWE+ResNext & 4.0 & 28.6 & 54.8 & 65.6 \\ 
% \bottomrule
% \end{tabular}
% \end{center}
% 
% \end{table}

\begin{table*}[t]
\caption{Performance of the extensions of our method on \textsc{Recipe1M} image-to-textual-recipe task and the previous state-of-the-art method, ACME \cite{wang2019learning}. The best results are shown in bold, and are statistically significant. While we report the results for both 1K and 10K test sample size, we only rely on 10K values for our analysis since 1K results are too noisy \cite{Carvalho2018}. 
}
\label{table:results2}
\begin{center}
\begin{tabular}{c l rrrr}  
\toprule
Size of test set & Method & medR $\downarrow$ & R@1 $\uparrow$ & R@5 $\uparrow$ & R@10 $\uparrow$ \\
\midrule
\multirow{5}{*}{1K} & (Previous SoTA) ACME \cite{wang2019learning} & 1.0 & 51.8 & 80.2 & 87.5  \\
& (Ours) CkNN+AWE+ResNet & 2.0 & 45.7 & 75.9 & 84.2 \\
& (Ours) CkNN+AWE+ResNext & 1.3 & 50.5 & 79.5 & 86.7 \\
& (Ours) Triplet+AWE+ResNet & 1.0 & 55.9 & 82.4 & 88.7 \\
& \textbf{(Ours) Triplet+AWE+ResNext} & \textbf{1.0} & \textbf{60.2} & \textbf{84.0} & \textbf{89.7}\\
\midrule
\multirow{5}{*}{10K} & (Previous SoTA) ACME \cite{wang2019learning} & 6.7 & 22.9 & 46.8 & 57.9 \\
& (Ours) CkNN+AWE+ResNet & 9.1 & 19.1 & 41.3 &	52.5 \\
& (Ours) CkNN+AWE+ResNext & 6.8 & 22.9 & 46.9 & 58.0 \\
& (Ours) Triplet+AWE+ResNet & 5.0 & 26.5 & 51.8 & 62.6 \\
% & (Ours) $^\dag$Triplet+AWE+ResNet & 5.9 & 24.4 & 49.4 & 60.5 \\
& \textbf{(Ours) Triplet+AWE+ResNext } & \textbf{4.0} & \textbf{30.0} & \textbf{56.5} & \textbf{67.0} \\
% & (Ours) $^\dag$Triplet+AWE+ResNext & 4.0 & 28.6 & 54.8 & 65.6 \\ 
\bottomrule
\end{tabular}
\end{center}
\end{table*}

In this section we consider generalisations on top of the base CkNN model. While ACME* image and text embeddings combination outperforms all others according to Table \ref{table:results}, we note that ResNet-Encoder and AWE-Encoder score is very close with 10K-R@1 of 17.9 and 17.4 respectively. In fact, CkNN combined with the embeddings precomputed using these encoders (\textit{CkNN+AWE+ResNet} in Table \ref{table:results1}) provides a strong cross-modal recipe retrieval baseline that improves upon a third-best published result \cite{Zhu_2019_CVPR} In this section, we show how to improve \textit{CkNN+AWE+ResNet} baseline to achieve new SoTA, summarizing our results in Table \ref{table:results2}.

\subsubsection{Triplet Loss Alignment}

Although we showed that CkNN is a suitable choice for model comparison and building cross-modal retrieval baselines, there is still scope for improving alignment. As observed in Section \ref{section:comparison}, direct search through a jointly learned embedding space can surpass the results of our CkNN approach for ACME* and Pic2Recipe encoders. Thus, we also train a standard triplet loss alignment module to create a joint embedding space on top of the precomputed image and text embeddings.

\begin{figure*}[t]
    \centering
    \includegraphics[width=0.7\textwidth]{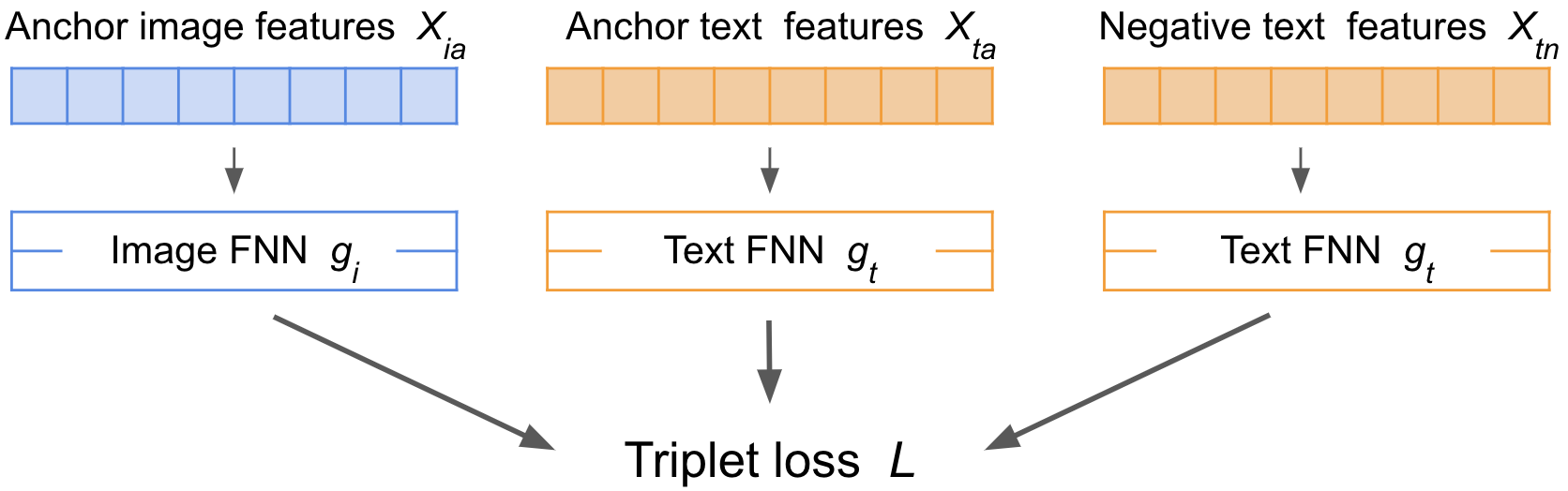}
    \caption{Projecting the precomputed image and textual features in the same common space using triplet loss. $X_{ia}$ has $D_i$ dimensions, $X_{ta}$ and $X_{tn}$ have $D_t$ dimensions}
    \label{fig:triplet}
\end{figure*}

In particular, we jointly train two feed-forward neural networks (FNN) with one hidden layer, dropout and batch normalization: one for image ($g_i$) and another for textual ($g_t$) features with triplet loss. Architectures of both neural networks are identical, and output feature size is $D=1024$. This pipeline is depicted in \figurename \ref{fig:triplet}.

Each triplet consists of one feature embedding as an anchor point in image modality and a positive and negative feature embeddings from text modality. The positive instances are the different modalities of the same recipe, $X_{ia}$ for image and $X_{ta}$ for textual features. We use online negative instance mining \cite{Schroff2015facenet} to choose the negative instance $X_{tn}$ as the closest text instance to the anchor point selected from other recipes in the mini-batch. The objective $\mathcal{L}$ is given by Eq. (\ref{eq:triplet}).

\begin{equation}
\label{eq:triplet}
\begin{split}
\mathcal{L} = \max(0, & d(g_i(X_{ia}), g_t(X_{ta})) - \\ 
 & d(g_i(X_{ia}), g_t(X_{tn})) + \gamma),
\end{split}
\end{equation}

\noindent where $d$ is the cosine distance between two vectors and $\gamma$ is the margin.

The model was trained on 238K recipes from the \textsc{Recipe1M} training set, with the margin $\gamma$ manually tuned to be 0.3 on 10K-R@1 metric on a validation set. We used Adam optimizer \cite{kingma2014adam}, a batch size of 256, initial learning rate of 0.002, and applied alternating optimization \cite{Salvadora} to aid convergence. Training takes only 25 seconds per epoch on a Tesla M60 GPU. The hyperparameters were tuned on a validation set.

This model boosts 10K-R@1 metric to 26.5 on \textsc{Recipe1M}, yielding a new SoTA result (\textit{Triplet+AWE+ResNet} in Table \ref{table:results2}). We emphasize that these SoTA results were achieved by applying a small alignment module on top of features precomputed from an independently trained ResNet-50 image encoder and an average word embeddings textual recipe encoder. This is in contrast to the complexity of the previous SoTA approach, which jointly trained image and text models using GANs; an adversarial alignment module; a novel hard negative mining strategy; translation consistency losses; classification losses; and multiple bidirectional LSTMs on top of skip-thought vectors and dedicated ingredient embeddings for textual recipe representation \cite{wang2019learning}. This suggests that the retrieval metrics on the \textsc{Recipe1M} dataset still have a lot of room for improvement, and we expect large gains to be achieved with advanced end-to-end methods in future work.

\subsubsection{Increasing the Capacity of the Image Encoder}

Since our ResNet-Encoder is trained using a classification objective, it is an obvious extension to replace \mbox{ResNet-50} with ResNext-101 \cite{Xie2016} architecture (ResNext-Encoder), which performs better on standard classification benchmarks such as ImageNet \cite{imagenet_cvpr09}.

When we use ResNext-Encoder, 10K-R@1 metric for CkNN reaches 22.9 (\textit{CkNN+AWE+ResNext}), matching previous SoTA results from \cite{wang2019learning}, and with triplet loss (\textit{Triplet+AWE+ResNext}) 10K-R@1 is boosted to 30.0, which further improves on previous SoTA by a large margin with the relative change of 30\%. It remains to be seen to what extent the existing methods would benefit from other image architectures.

\subsubsection{Ablation Study: Impact of Training Data Utilization}

\begin{table*}[t]
\caption{Ablation study of the new SoTA results. $^\dag$ symbol next to the model means using only training data available to ACME \cite{wang2019learning} method after preprocessing and selecting recipes paired with images. 
We only report on 10K test set since 1K results are too noisy \cite{Carvalho2018}.}
\label{table:ablation}
\begin{center}
\begin{tabular}{l rrrr}  
\toprule
Method & medR $\downarrow$ & R@1 $\uparrow$ & R@5 $\uparrow$ & R@10 $\uparrow$ \\
\midrule
(Previous SoTA) ACME \cite{wang2019learning} & 6.7 & 22.9 & 46.8 & 57.9 \\
(Ours) $^\dag$Triplet+AWE+ResNet & 5.9 & 24.4 & 49.4 & 60.5 \\ 
(Ours) Triplet+AWE+ResNet & 5.0 & 26.5 & 51.8 & 62.6 \\
(Ours) $^\dag$Triplet+AWE+ResNext & 4.0 & 28.6 & 54.8 & 65.6 \\ 
\textbf{(Ours) Triplet+AWE+ResNext} & \textbf{4.0} & \textbf{30.0} & \textbf{56.5} & \textbf{67.0} \\
\bottomrule
\end{tabular}
%}
\end{center}
\end{table*}

One benefit of training text and image models separately as described in Section \ref{section:awe_encoder} and \ref{section:recipe1m_classifier}, is that one can make use of training data which was filtered out by the existing methods' preprocessing pipeline \cite{Salvadora,wang2019learning}. Indeed, since the setup for training AWE-Encoder does not require any image-to-textual-recipe pairs and is entirely self-supervised, we are able to train it on 680K textual recipes from the \textsc{Recipe1M} training set, as opposed to only 240K recipes with images filtered by ACME for joint training \cite{wang2019learning}. Similarly, we utilize 280K recipes with images for ResNet-Encoder/ResNext-Encoder compared to 240K filtered by ACME \cite{wang2019learning}. It should be noted, however, that ingredient and instruction embeddings from Pic2Recipe and ACME were also trained on the full 1M dataset \cite{Salvadora}.

To see by how much our best models' performance was improved by better training data utilization, we train our encoders only on the subset of images and recipes used by ACME \cite{wang2019learning} for joint training. Triplet loss model has already been using the same subset. We observe that 10K-R@1 metric dropped from 30.0 to 28.6 for ResNext, and from 26.5 to 24.4 for ResNet. While this difference is statistically significant, all the models in the ablation study still improve on previous SoTA (\textit{$^\dag$Triplet+AWE+ResNet} and \textit{$^\dag$Triplet+AWE+ResNext} in Table \ref{table:ablation}).

% \subsubsection{Triplet Loss on \textsc{Politics} and \textsc{GoodNews}}
% % 
% We extend the triplet-based generalisation of CkNN to the \textsc{Politics} and \textsc{GoodNews} datasets in this section. The results are shown in Table \ref{table:triplet_politics_goodnews}, and SoTA performance is denoted with bold font. We can see from this image that introducing triplet mining on top of CkNN module reaches and exceeds SoTA performance. Although we have set new SoTA performance on these, we recognise that our improvements are models and are much more comprehensive on \textsc{Recipe1M}. However, the fact that the application of minor changes to CkNN empowers it to reach SoTA on all three datasets is a promising endorsement of our proposed model. 

% \begin{table}[]
%     \centering
%     \begin{tabular}{llrr}
%         \toprule
%                  &       & T$\rightarrow$I & I$\rightarrow$T  \\
%          Dataset & Model & 10K-R@1 & 10K-R@1  \\
%          \midrule
%          \textsc{Politics} & Triplet+AWE+ResNet       & \textbf{64.8} & \textbf{64.3} \\
%          \textsc{Politics} & Triplet+AWE+ResNext      & & \\
%          \midrule
%          \textsc{GoodNews} & Triplet+FastText+ResNet  & 88.2 & 88.3 \\
%          \textsc{GoodNews} & Triplet+FastText+ResNext & \textbf{88.6} & 88.5 \\
%          \bottomrule
%     \end{tabular}
%     \caption{Results of triplet-based models on \textsc{Politics} and \textsc{GoodNews} datasets. Bold font implies SoTA performance.}
%     \label{table:triplet_politics_goodnews}
% \end{table}

\subsection{Performance on \textsc{Politics} and \textsc{GoodNews}}
\label{section:other}

To understand how well our approach generalizes to other (non-cooking) domains, we apply our method to other large-scale publicly available datasets. We hypothesize that our method will be effective for challenging cross-modal tasks where many of the approaches developed for benchmarks such as Flickr30k will not be applicable, and thus strong baselines might be of value. We identify two such datasets. \textsc{Politics} \cite{thomas2019predicting} consists of 246K political articles paired with images after preprocessing. \textsc{GoodNews} \cite{biten2019good} features 466K images from news articles paired with captions. We split \textsc{GoodNews} and \textsc{Politics} datasets into training, validation and test sets with 80-10-10 ratio following Thomas \textit{et al} \cite{thomas2020preserving}. As for \textsc{Recipe1M}, we focused on the Recall@1 metric on image-to-document retrieval: for each image, we randomly sample 4 candidate documents from the evaluation set, and report the proportion of query images for which the matching document was ranked higher than all the candidate documents as in prior work \cite{thomas2020preserving}.

We then apply our CkNN baseline to these datasets, analogously to how it was done for \textsc{Recipe1M} in Section \ref{section:competitive}. Specifically, for \textsc{Politics} dataset, we have used the first two sentences of the text document as a title following \cite{thomas2020preserving}, and extracted $C_i=C_t=2527$ labels for training AWE-Encoder and ResNet-Encoder. For \textsc{GoodNews} dataset we have used the captions to extract the $C_i=2120$ labels for ResNet-Encoder, and resorted to using FastText \cite{bojanowski2016enriching} for embedding the words of the short captions as discussed in Section \ref{section:awe_encoder}. A sample of extracted labels and documents were manually validated to verify relevance.

For both datasets, we used CkNN to align the modalities. We used exactly the same hyperparameters for CkNN, ResNet-Encoder and AWE-Encoder as we did for \textsc{Recipe1M} in Section \ref{section:competitive}. We report CkNN performance on \textsc{Politics} and \textsc{GoodNews} datasets in  Table \ref{table:politics}. We further report the performance of our models on \textsc{Politics} and \textsc{GoodNews} after applying the improvements mentioned in Section \ref{section:sota} to the respective CkNN baselines described in the current section. For \textsc{Politics}, our method achieves Recall@1 of 64.8 on the image-to-text task. For \textsc{GoodNews}, it achieves Recall@1 of 88.6 on the image-to-text task. The results in each case are on par with the best published method (see Table \ref{table:politics} and \cite{thomas2020preserving}). 

Although the performance differences between our work and that of \cite{thomas2020preserving} is not statistically significant, our approach, which from one perspective may be viewed as a baseline, is highly competitive and matches the leading methods on these datasets and tasks. We observe that the existing methods for cross-modal recipe retrieval are specialized to and evaluated on only a single Recipe1M dataset, whereas our method is also competitive on \textsc{Politics} and \textsc{GoodNews} without any modifications, and while using exactly the same hyperparameters.

\begin{table*}[t]
\caption{Recall@1 performance of our baselines and other approaches \cite{thomas2020preserving} on \textsc{Politics} and \textsc{GoodNews} for the 5-way image-to-text retrieval task. The best results within the level of statistical significance are in bold.}
\label{table:politics}
\begin{center}
\begin{tabular}{l rr}  
\toprule
Method & \textsc{Politics} & \textsc{GoodNews} \\
\midrule
Trip+NP+Sym & 47.4 & 72.0 \\
PVSE & 62.5 & 87.2 \\
Ang+NP+Sym & 62.7 & 87.0 \\
\textbf{(Previous SoTA) SN} & \textbf{64.7} & \textbf{88.5} \\
(Ours) CkNN+AWE+ResNet & 60.5 & 83.8 \\ %79.7
\textbf{(Ours) Triplet+AWE+ResNext} & \textbf{64.8} & \textbf{88.6} \\
\bottomrule
\end{tabular}
\end{center}

\end{table*}

\subsection{Text-To-Image Retrieval}
\label{section:image-to-text}
Although in this paper we only focus on image-to-text retrieval tasks for clarity and the ease of analysis, we also report our best model's performances on the mirror text-to-image task for completeness. On \textsc{Recipe1M}, the proposed method achieves medR of 4.0, 10K-R@1 of 30.5, 10K-R@5 of 56.3 and 10K-R@10 of 66.6. Similar to the image-to-text task, this a large improvement over the previous SoTA results from \cite{wang2019learning}. Text-to-image retrieval is also very good on \textsc{Politics} and \textsc{GoodNews} with Recall@1 of 64.9 and 88.5 respectively which is on par with SoTA performance on these datasets \cite{thomas2020preserving}. 
%Generally speaking, we observed approximately symmetric results on \textsc{Politics} and \textsc{GoodNews}, mirroring the results of \cite{thomas2020preserving}. %Approximately equivalent performance is achieved on the symmetric tasks with \textsc{Politics} and \textsc{GoodNews} \cite{thomas2020preserving}.

\section{Conclusion} \label{section:conclusions}

% Our work presents that Nearest-Neighbour-based approaches allow for easy creation of strong baselines for non-literal cross-modal retrieval. We further demonstrate that all the existing methods for cross-modal recipe retrieval could be outperformed with unsophisticated approaches, echoing the sentiment of \citet{Dacrema:2019aa} from neural recommendation. We hope that our comparison framework would help researchers to improve individual modules of advanced jointly trained methods, while the presented baseline methods and SoTA results would move the goalposts to aid further progress.

We conclude this work by highlighting two key takeaway messages. First, this paper provides strong evidence that nearest neighbours, when incorporated according to our methodology, offers an straightforward, yet broadly performant, baseline on a cross-modal recipe retrieval task. The test performance of our proposed model is competitive with SoTA evaluation, even though we deliberately avoided `advanced' modelling techniques. The benefit of this is efficient, stable and reliable solutions with relatively low computational overhead in training and evaluation. Although these baseline results do not best SoTA, they come reasonably close to it. Secondly, we show that our approach greatly simplifies model exploration and model comparison. We demonstrate in a case study how to compare various models' individual components using our method and use the insights from the analysis to significantly advance SoTA on the definitive cross-modal recipe retrieval dataset, \textsc{Recipe1M}, with straightforward modifications of our baseline. The resulting model has much less complexity than other methods tailored for \textsc{Recipe1M}, and it generalizes well enough to match best published results on two other challenging datasets for cross-modal retrieval. We believe that our approach fills a growing need for strong baselines and a systematic comparison framework in cross-modal recipe retrieval and hope that it will facilitate further progress in this space as well as other cross-modal domains.

\ifCLASSOPTIONcaptionsoff
  \newpage
\fi



%\bibliographystyle{IEEEtran}



% \end{thebibliography}

\bibliographystyle{IEEEtran}
\bibliography{refs}

\begin{thebibliography}{10}
\providecommand{\url}[1]{#1}
\csname url@samestyle\endcsname
\providecommand{\newblock}{\relax}
\providecommand{\bibinfo}[2]{#2}
\providecommand{\BIBentrySTDinterwordspacing}{\spaceskip=0pt\relax}
\providecommand{\BIBentryALTinterwordstretchfactor}{4}
\providecommand{\BIBentryALTinterwordspacing}{\spaceskip=\fontdimen2\font plus
\BIBentryALTinterwordstretchfactor\fontdimen3\font minus
  \fontdimen4\font\relax}
\providecommand{\BIBforeignlanguage}[2]{{%
\expandafter\ifx\csname l@#1\endcsname\relax
\typeout{** WARNING: IEEEtran.bst: No hyphenation pattern has been}%
\typeout{** loaded for the language `#1'. Using the pattern for}%
\typeout{** the default language instead.}%
\else
\language=\csname l@#1\endcsname
\fi
#2}}
\providecommand{\BIBdecl}{\relax}
\BIBdecl

\bibitem{Marin2018}
\BIBentryALTinterwordspacing
J.~Mar{\'{\i}}n, A.~Biswas, F.~Ofli, N.~Hynes, A.~Salvador, Y.~Aytar, I.~Weber,
  and A.~Torralba, ``Recipe1m: {A} dataset for learning cross-modal embeddings
  for cooking recipes and food images,'' \emph{arXiv preprint
  arXiv:1810.06553}, vol. abs/1810.06553, 2018. [Online]. Available:
  \url{http://arxiv.org/abs/1810.06553}
\BIBentrySTDinterwordspacing

\bibitem{Myers2015}
\BIBentryALTinterwordspacing
A.~Myers, N.~Johnston, V.~Rathod, A.~Korattikara, A.~Gorban, N.~Silberman,
  S.~Guadarrama, G.~Papandreou, J.~Huang, and K.~Murphy, ``{Im2Calories:
  Towards an Automated Mobile Vision Food Diary},'' in \emph{2015 IEEE
  International Conference on Computer Vision (ICCV)}.\hskip 1em plus 0.5em
  minus 0.4em\relax IEEE, 2015, pp. 1233--1241. [Online]. Available:
  \url{http://ieeexplore.ieee.org/document/7410503/}
\BIBentrySTDinterwordspacing

\bibitem{Freyne}
J.~Freyne and S.~Berkovsky, ``Recommending food: Reasoning on recipes and
  ingredients,'' in \emph{User Modeling, Adaptation, and
  Personalization}.\hskip 1em plus 0.5em minus 0.4em\relax Berlin, Heidelberg:
  Springer Berlin Heidelberg, 2010, pp. 381--386.

\bibitem{Salvadora}
A.~{Salvador}, N.~{Hynes}, Y.~{Aytar}, J.~{Marin}, F.~{Ofli}, I.~{Weber}, and
  A.~{Torralba}, ``Learning cross-modal embeddings for cooking recipes and food
  images,'' in \emph{2017 IEEE Conference on Computer Vision and Pattern
  Recognition (CVPR)}, July 2017, pp. 3068--3076.

\bibitem{cca}
\BIBentryALTinterwordspacing
H.~Hotelling, ``{Relations between two sets of variates},'' \emph{Biometrika},
  vol.~28, no. 3-4, pp. 321--377, 12 1936. [Online]. Available:
  \url{https://doi.org/10.1093/biomet/28.3-4.321}
\BIBentrySTDinterwordspacing

\bibitem{Chen2017crossmodal}
\BIBentryALTinterwordspacing
J.-j. Chen, C.-W. Ngo, and T.-S. Chua, ``Cross-modal recipe retrieval with rich
  food attributes,'' in \emph{Proceedings of the 25th ACM International
  Conference on Multimedia}, ser. MM '17.\hskip 1em plus 0.5em minus
  0.4em\relax New York, NY, USA: ACM, 2017, pp. 1771--1779. [Online].
  Available: \url{http://doi.acm.org/10.1145/3123266.3123428}
\BIBentrySTDinterwordspacing

\bibitem{Chen2018crossmodal}
\BIBentryALTinterwordspacing
J.-J. Chen, C.-W. Ngo, F.-L. Feng, and T.-S. Chua, ``Deep understanding of
  cooking procedure for cross-modal recipe retrieval,'' in \emph{Proceedings of
  the 26th ACM International Conference on Multimedia}, ser. MM '18.\hskip 1em
  plus 0.5em minus 0.4em\relax New York, NY, USA: ACM, 2018, pp. 1020--1028.
  [Online]. Available: \url{http://doi.acm.org/10.1145/3240508.3240627}
\BIBentrySTDinterwordspacing

\bibitem{Carvalho2018}
\BIBentryALTinterwordspacing
M.~Carvalho, R.~Cad\`{e}ne, D.~Picard, L.~Soulier, N.~Thome, and M.~Cord,
  ``Cross-modal retrieval in the cooking context: Learning semantic text-image
  embeddings,'' in \emph{The 41st International ACM SIGIR Conference on
  Research \&\#38; Development in Information Retrieval}, ser. SIGIR '18.\hskip
  1em plus 0.5em minus 0.4em\relax New York, NY, USA: ACM, 2018, pp. 35--44.
  [Online]. Available: \url{http://doi.acm.org/10.1145/3209978.3210036}
\BIBentrySTDinterwordspacing

\bibitem{Zhu_2019_CVPR}
B.~Zhu, C.-W. Ngo, J.~Chen, and Y.~Hao, ``{R2GAN}: Cross-modal recipe retrieval
  with generative adversarial network,'' in \emph{The IEEE Conference on
  Computer Vision and Pattern Recognition (CVPR)}, June 2019.

\bibitem{wang2019learning}
H.~Wang, D.~Sahoo, C.~Liu, E.-p. Lim, and S.~C.~H. Hoi, ``Learning cross-modal
  embeddings with adversarial networks for cooking recipes and food images,''
  in \emph{Proceedings of the IEEE Conference on Computer Vision and Pattern
  Recognition}, 2019, pp. 11\,572--11\,581.

\bibitem{thomas2019predicting}
C.~Thomas and A.~Kovashka, ``Predicting the politics of an image using webly
  supervised data,'' in \emph{Advances in Neural Information Processing
  Systems}, 2019, pp. 3630--3642.

\bibitem{biten2019good}
A.~F. Biten, L.~Gomez, M.~Rusinol, and D.~Karatzas, ``Good news, everyone!
  context driven entity-aware captioning for news images,'' in
  \emph{Proceedings of the IEEE Conference on Computer Vision and Pattern
  Recognition}, 2019, pp. 12\,466--12\,475.

\bibitem{thomas2020preserving}
C.~Thomas and A.~Kovashka, ``Preserving semantic neighborhoods for robust
  cross-modal retrieval,'' in \emph{Proceedings of the European Conference on
  Computer Vision (ECCV)}, August 2020.

\bibitem{plummer2015flickr30k}
B.~A. Plummer, L.~Wang, C.~M. Cervantes, J.~C. Caicedo, J.~Hockenmaier, and
  S.~Lazebnik, ``Flickr30k entities: Collecting region-to-phrase
  correspondences for richer image-to-sentence models,'' in \emph{Proceedings
  of the IEEE international conference on computer vision}, 2015, pp.
  2641--2649.

\bibitem{lin2014microsoft}
T.-Y. Lin, M.~Maire, S.~Belongie, J.~Hays, P.~Perona, D.~Ramanan,
  P.~Doll{\'a}r, and C.~L. Zitnick, ``Microsoft coco: Common objects in
  context,'' in \emph{European conference on computer vision}.\hskip 1em plus
  0.5em minus 0.4em\relax Springer, 2014, pp. 740--755.

\bibitem{jiang2017}
Q.~{Jiang} and W.~{Li}, ``Deep cross-modal hashing,'' in \emph{2017 IEEE
  Conference on Computer Vision and Pattern Recognition (CVPR)}, 2017, pp.
  3270--3278.

\bibitem{tkde_wang2020}
Y.~{Wang}, X.~{Luo}, L.~{Nie}, J.~{Song}, W.~{Zhang}, and X.~{Xu}, ``Batch: A
  scalable asymmetric discrete cross-modal hashing,'' \emph{IEEE Transactions
  on Knowledge and Data Engineering}, pp. 1--1, 2020.

\bibitem{Feng2014}
\BIBentryALTinterwordspacing
F.~Feng, X.~Wang, and R.~Li, ``Cross-modal retrieval with correspondence
  autoencoder,'' in \emph{Proceedings of the 22Nd ACM International Conference
  on Multimedia}, ser. MM '14.\hskip 1em plus 0.5em minus 0.4em\relax New York,
  NY, USA: ACM, 2014, pp. 7--16. [Online]. Available:
  \url{http://doi.acm.org/10.1145/2647868.2654902}
\BIBentrySTDinterwordspacing

\bibitem{Ngiam2011}
\BIBentryALTinterwordspacing
J.~Ngiam, A.~Khosla, M.~Kim, J.~Nam, H.~Lee, and A.~Y. Ng, ``Multimodal deep
  learning,'' in \emph{Proceedings of the 28th International Conference on
  International Conference on Machine Learning}, ser. ICML'11.\hskip 1em plus
  0.5em minus 0.4em\relax USA: Omnipress, 2011, pp. 689--696. [Online].
  Available: \url{http://dl.acm.org/citation.cfm?id=3104482.3104569}
\BIBentrySTDinterwordspacing

\bibitem{tkde_tu2020}
R.~{Tu}, X.~{Mao}, B.~{Ma}, Y.~{Hu}, T.~{Yan}, W.~{Wei}, and H.~{Huang}, ``Deep
  cross-modal hashing with hashing functions and unified hash codes jointly
  learning,'' \emph{IEEE Transactions on Knowledge and Data Engineering}, pp.
  1--1, 2020.

\bibitem{Peng2017CCL}
Y.~Peng, J.~Qi, X.~Huang, and Y.~Yuan, ``{CCL}: Cross-modal correlation
  learning with multigrained fusion by hierarchical network,'' \emph{IEEE
  Transactions on Multimedia}, vol.~PP, 04 2017.

\bibitem{Wang2017}
\BIBentryALTinterwordspacing
B.~Wang, Y.~Yang, X.~Xu, A.~Hanjalic, and H.~T. Shen, ``Adversarial cross-modal
  retrieval,'' in \emph{Proceedings of the 25th ACM International Conference on
  Multimedia}, ser. MM '17.\hskip 1em plus 0.5em minus 0.4em\relax New York,
  NY, USA: ACM, 2017, pp. 154--162. [Online]. Available:
  \url{http://doi.acm.org/10.1145/3123266.3123326}
\BIBentrySTDinterwordspacing

\bibitem{tkde_shen2020}
H.~T. {Shen}, L.~{Liu}, Y.~{Yang}, X.~{Xu}, Z.~{Huang}, F.~{Shen}, and
  R.~{Hong}, ``Exploiting subspace relation in semantic labels for cross-modal
  hashing,'' \emph{IEEE Transactions on Knowledge and Data Engineering}, pp.
  1--1, 2020.

\bibitem{gan}
\BIBentryALTinterwordspacing
I.~J. Goodfellow, J.~Pouget-Abadie, M.~Mirza, B.~Xu, D.~Warde-Farley, S.~Ozair,
  A.~Courville, and Y.~Bengio, ``Generative adversarial nets,'' in
  \emph{Proceedings of the 27th International Conference on Neural Information
  Processing Systems - Volume 2}, ser. NIPS'14.\hskip 1em plus 0.5em minus
  0.4em\relax Cambridge, MA, USA: MIT Press, 2014, pp. 2672--2680. [Online].
  Available: \url{http://dl.acm.org/citation.cfm?id=2969033.2969125}
\BIBentrySTDinterwordspacing

\bibitem{Peng2017}
Y.~Peng, J.~Qi, and Y.~Yuan, ``Cm-gans: Cross-modal generative adversarial
  networks for common representation learning,'' \emph{ACM Transactions on
  Multimedia Computing, Communications, and Applications}, vol.~15, 10 2017.

\bibitem{Lee2018}
K.-H. Lee, X.~Chen, G.~Hua, H.~Hu, and X.~He, ``Stacked cross attention for
  image-text matching,'' in \emph{The European Conference on Computer Vision
  (ECCV)}, September 2018.

\bibitem{wang2019camp}
Z.~Wang, X.~Liu, H.~Li, L.~Sheng, J.~Yan, X.~Wang, and J.~Shao, ``Camp:
  Cross-modal adaptive message passing for text-image retrieval,'' in
  \emph{Proceedings of the IEEE International Conference on Computer Vision},
  2019, pp. 5764--5773.

\bibitem{resnet50}
K.~{He}, X.~{Zhang}, S.~{Ren}, and J.~{Sun}, ``Deep residual learning for image
  recognition,'' in \emph{2016 IEEE Conference on Computer Vision and Pattern
  Recognition (CVPR)}, June 2016, pp. 770--778.

\bibitem{hochreiter1997}
\BIBentryALTinterwordspacing
S.~Hochreiter and J.~Schmidhuber, ``Long short-term memory,'' \emph{Neural
  Comput.}, vol.~9, no.~8, pp. 1735--1780, Nov. 1997. [Online]. Available:
  \url{http://dx.doi.org/10.1162/neco.1997.9.8.1735}
\BIBentrySTDinterwordspacing

\bibitem{mikolov2013}
\BIBentryALTinterwordspacing
T.~Mikolov, I.~Sutskever, K.~Chen, G.~Corrado, and J.~Dean, ``Distributed
  representations of words and phrases and their compositionality,'' in
  \emph{Proceedings of the 26th International Conference on Neural Information
  Processing Systems - Volume 2}, ser. NIPS'13.\hskip 1em plus 0.5em minus
  0.4em\relax USA: Curran Associates Inc., 2013, pp. 3111--3119. [Online].
  Available: \url{http://dl.acm.org/citation.cfm?id=2999792.2999959}
\BIBentrySTDinterwordspacing

\bibitem{kiros2015}
\BIBentryALTinterwordspacing
R.~Kiros, Y.~Zhu, R.~Salakhutdinov, R.~S. Zemel, A.~Torralba, R.~Urtasun, and
  S.~Fidler, ``Skip-thought vectors,'' in \emph{Proceedings of the 28th
  International Conference on Neural Information Processing Systems - Volume
  2}, ser. NIPS'15.\hskip 1em plus 0.5em minus 0.4em\relax Cambridge, MA, USA:
  MIT Press, 2015, pp. 3294--3302. [Online]. Available:
  \url{http://dl.acm.org/citation.cfm?id=2969442.2969607}
\BIBentrySTDinterwordspacing

\bibitem{chen2017crossmodal_b}
J.~Chen, L.~Pang, and C.-W. Ngo, ``Cross-modal recipe retrieval: How to cook
  this dish?'' in \emph{MultiMedia Modeling}.\hskip 1em plus 0.5em minus
  0.4em\relax Cham: Springer International Publishing, 2017, pp. 588--600.

\bibitem{Fu_2020_CVPR}
H.~Fu, R.~Wu, C.~Liu, and J.~Sun, ``Mcen: Bridging cross-modal gap between
  cooking recipes and dish images with latent variable model,'' in
  \emph{Proceedings of the IEEE/CVF Conference on Computer Vision and Pattern
  Recognition (CVPR)}, June 2020.

\bibitem{Schroff2015}
F.~Schroff, D.~Kalenichenko, and J.~Philbin, ``{FaceNet}: A unified embedding
  for face recognition and clustering,'' in \emph{2015 IEEE Conference on
  Computer Vision and Pattern Recognition (CVPR)}, June 2015, pp. 815--823.

\bibitem{Bossard}
L.~Bossard, M.~Guillaumin, and L.~Van~Gool, ``Food-101 -- mining discriminative
  components with random forests,'' in \emph{European Conference on Computer
  Vision}, 2014.

\bibitem{Chen2017}
\BIBentryALTinterwordspacing
X.~Chen, H.~Zhou, and L.~Diao, ``{ChineseFoodNet}: {A} large-scale image
  dataset for {C}hinese food recognition,'' \emph{arXiv preprint
  arXiv:1705.02743}, 2017. [Online]. Available:
  \url{http://arxiv.org/abs/1705.02743}
\BIBentrySTDinterwordspacing

\bibitem{Hassannejad2016}
\BIBentryALTinterwordspacing
H.~Hassannejad, G.~Matrella, P.~Ciampolini, I.~De~Munari, M.~Mordonini, and
  S.~Cagnoni, ``Food image recognition using very deep convolutional
  networks,'' in \emph{Proceedings of the 2Nd International Workshop on
  Multimedia Assisted Dietary Management}, ser. MADiMa '16.\hskip 1em plus
  0.5em minus 0.4em\relax New York, NY, USA: ACM, 2016, pp. 41--49. [Online].
  Available: \url{http://doi.acm.org/10.1145/2986035.2986042}
\BIBentrySTDinterwordspacing

\bibitem{Singla2016}
\BIBentryALTinterwordspacing
A.~Singla, L.~Yuan, and T.~Ebrahimi, ``Food/non-food image classification and
  food categorization using pre-trained {GoogLeNet} model,'' in
  \emph{Proceedings of the 2Nd International Workshop on Multimedia Assisted
  Dietary Management}, ser. MADiMa '16.\hskip 1em plus 0.5em minus 0.4em\relax
  New York, NY, USA: ACM, 2016, pp. 3--11. [Online]. Available:
  \url{http://doi.acm.org/10.1145/2986035.2986039}
\BIBentrySTDinterwordspacing

\bibitem{food101sota}
N.~{Martinel}, G.~L. {Foresti}, and C.~{Micheloni}, ``Wide-slice residual
  networks for food recognition,'' in \emph{2018 IEEE Winter Conference on
  Applications of Computer Vision (WACV)}, March 2018, pp. 567--576.

\bibitem{sato-etal-2016-japanese}
\BIBentryALTinterwordspacing
T.~Sato, J.~Harashima, and M.~Komachi, ``{J}apanese-{E}nglish machine
  translation of recipe texts,'' in \emph{Proceedings of the 3rd Workshop on
  {A}sian Translation ({WAT}2016)}.\hskip 1em plus 0.5em minus 0.4em\relax
  Osaka, Japan: The COLING 2016 Organizing Committee, Dec. 2016, pp. 58--67.
  [Online]. Available: \url{https://www.aclweb.org/anthology/W16-4603}
\BIBentrySTDinterwordspacing

\bibitem{Donghyeon2019}
\BIBentryALTinterwordspacing
D.~Park, K.~Kim, Y.~Park, J.~Shin, and J.~Kang, ``Kitchenette: Predicting and
  ranking food ingredient pairings using siamese neural network,'' in
  \emph{Proceedings of the Twenty-Eighth International Joint Conference on
  Artificial Intelligence, {IJCAI-19}}.\hskip 1em plus 0.5em minus 0.4em\relax
  International Joint Conferences on Artificial Intelligence Organization, 7
  2019, pp. 5930--5936. [Online]. Available:
  \url{https://doi.org/10.24963/ijcai.2019/822}
\BIBentrySTDinterwordspacing

\bibitem{Dacrema:2019aa}
M.~F. Dacrema \emph{et~al.}, ``{A}re {W}e {R}eally {M}aking {M}uch {P}rogress?
  {A} {W}orrying {A}nalysis of {R}ecent {N}eural {R}ecommendation
  {A}pproaches,'' Proceedings of the 13th ACM Conference on Recommender Systems
  (RecSys 2019), 2019.

\bibitem{Ondrej2007}
O.~{Chum}, J.~{Philbin}, J.~{Sivic}, M.~{Isard}, and A.~{Zisserman}, ``Total
  recall: Automatic query expansion with a generative feature model for object
  retrieval,'' in \emph{2007 IEEE 11th International Conference on Computer
  Vision}, Oct 2007, pp. 1--8.

\bibitem{Panu2009}
P.~{Turcot} and D.~{Lowe}, ``Better matching with fewer features: The selection
  of useful features in large database recognition problems,'' in \emph{2009
  IEEE 12th International Conference on Computer Vision Workshops, ICCV
  Workshops}, Sep. 2009, pp. 2109--2116.

\bibitem{Bert}
\BIBentryALTinterwordspacing
J.~Devlin, M.~Chang, K.~Lee, and K.~Toutanova, ``{BERT:} pre-training of deep
  bidirectional transformers for language understanding,'' \emph{arXiv preprint
  arxiv:1810.04805}, 2018. [Online]. Available:
  \url{http://arxiv.org/abs/1810.04805}
\BIBentrySTDinterwordspacing

\bibitem{Doersch2015}
\BIBentryALTinterwordspacing
C.~Doersch \emph{et~al.}, ``Unsupervised visual representation learning by
  context prediction,'' in \emph{Proceedings of the 2015 IEEE International
  Conference on Computer Vision (ICCV)}, ser. ICCV '15.\hskip 1em plus 0.5em
  minus 0.4em\relax Washington, DC, USA: IEEE Computer Society, 2015, pp.
  1422--1430. [Online]. Available:
  \url{http://dx.doi.org/10.1109/ICCV.2015.167}
\BIBentrySTDinterwordspacing

\bibitem{RetrievalCNN}
A.~Babenko, A.~Slesarev, A.~Chigorin, and V.~Lempitsky, ``Neural codes for
  image retrieval,'' in \emph{Computer Vision -- ECCV 2014}.\hskip 1em plus
  0.5em minus 0.4em\relax Cham: Springer International Publishing, 2014, pp.
  584--599.

\bibitem{Faster_R_CNN}
S.~Ren, K.~He, R.~Girshick, and J.~Sun, ``Faster {R-CNN}: Towards real-time
  object detection with region proposal networks,'' \emph{IEEE Transactions on
  Pattern Analysis and Machine Intelligence}, vol.~39, 06 2015.

\bibitem{Wieting2016}
J.~Wieting, M.~Bansal, K.~Gimpel, and K.~Livescu, ``Towards universal
  paraphrastic sentence embeddings,'' in \emph{International Conference on
  Learning Representations}, 2016.

\bibitem{Le2014}
Q.~Le and T.~Mikolov, ``Distributed representations of sentences and
  documents,'' \emph{31st International Conference on Machine Learning, ICML
  2014}, vol.~4, 05 2014.

\bibitem{Hill2016LearningDR}
F.~Hill, K.~Cho, and A.~Korhonen, ``Learning distributed representations of
  sentences from unlabelled data,'' in \emph{HLT-NAACL}, 2016.

\bibitem{tkde_guo2019}
S.~{Guo} and N.~{Yao}, ``Document vector extension for documents
  classification,'' \emph{IEEE Transactions on Knowledge and Data Engineering},
  pp. 1--1, 2019.

\bibitem{bojanowski2016enriching}
\BIBentryALTinterwordspacing
P.~Bojanowski, E.~Grave, A.~Joulin, and T.~Mikolov, ``Enriching word vectors
  with subword information,'' \emph{Transactions of the Association for
  Computational Linguistics}, vol.~5, pp. 135--146, 2017. [Online]. Available:
  \url{http://aclweb.org/anthology/Q17-1010}
\BIBentrySTDinterwordspacing

\bibitem{Deng2018ArcFaceAA}
J.~Deng, J.~Guo, and S.~Zafeiriou, ``{ArcFace}: Additive angular margin loss
  for deep face recognition,'' \emph{arXiv preprint arXiv:1801.07698}, 2018.

\bibitem{kingma2014adam}
D.~P. Kingma and J.~Ba, ``Adam: A method for stochastic optimization,''
  \emph{arXiv preprint arXiv:1412.6980}, 2014.

\bibitem{imagenet_cvpr09}
J.~Deng, W.~Dong, R.~Socher, L.-J. Li, K.~Li, and L.~Fei-Fei, ``{ImageNet: A
  Large-Scale Hierarchical Image Database},'' in \emph{CVPR09}, 2009.

\bibitem{Schroff2015facenet}
F.~Schroff, D.~Kalenichenko, and J.~Philbin, ``Facenet: A unified embedding for
  face recognition and clustering,'' in \emph{Proceedings of the IEEE
  conference on computer vision and pattern recognition}, 2015, pp. 815--823.

\bibitem{Xie2016}
S.~Xie, R.~Girshick, P.~Dollár, Z.~Tu, and K.~He, ``Aggregated residual
  transformations for deep neural networks,'' \emph{arXiv preprint
  arXiv:1611.05431}, 2016.

\end{thebibliography}


%\newpagece for a photo:


\vfill


\end{document}